\documentclass[sigconf]{acmart}
\usepackage{subcaption}
\usepackage{enumitem}
\usepackage{tabularx}
\usepackage{wrapfig}
\usepackage{lipsum}
\usepackage{amsmath}
\usepackage{multirow}
\usepackage{graphicx}
\usepackage{booktabs}
\usepackage{makecell}
\usepackage{algorithm}
\usepackage{algorithmic}
\usepackage{balance} 
\AtBeginDocument{%
  }

\copyrightyear{2026}
\acmYear{2026}
\setcopyright{cc}
\setcctype{by}
\acmConference[WSDM '26] {Proceedings of the Nineteenth ACM International Conference on Web Search and Data Mining}{February 22--26, 2026}{Boise, ID, USA.}
\acmBooktitle{Proceedings of the Nineteenth ACM International Conference on Web Search and Data Mining (WSDM '26), February 22--26, 2026, Boise, ID, USA}
\acmISBN{979-8-4007-2292-9/2026/02}
\acmDOI{10.1145/3773966.3777982}

\begin{document}

\title{Proficient Graph Neural Network Design by \\
Accumulating Knowledge on Large Language Models}

\author{Jialiang Wang}
\email{jwangic@connect.ust.hk}
\orcid{0009-0005-0850-0389}
\affiliation{%
  \institution{Hong Kong University of Science and Technology}
  \city{Hong Kong SAR}
  \country{China}
}

\author{Hanmo Liu}
\authornote{Also with Hong Kong University of Science and Technology (Guangzhou).}
\email{hliubm@connect.ust.hk}
\orcid{0000-0002-6471-0226}
\affiliation{%
  \institution{Hong Kong University of Science and Technology}
  \city{Hong Kong SAR}
  \country{China}
}

\author{Shimin Di}
\email{sdiaa@connect.ust.hk}
\orcid{0000-0002-7394-0082}
\affiliation{%
  \institution{Southeast University}
  \city{Nanjing}
  \country{China}
}

\author{Zhili Wang}
\email{zwangeo@connect.ust.hk}
\orcid{0000-0002-6737-2672}
\affiliation{%
 \institution{HoHuawei Hong Kong Research Center (HKRC)}
 \city{Hong Kong SAR}
 \country{China}
}

\author{Jiachuan Wang}
\authornote{Corresponding author.}
\email{wangjc@slis.tsukuba.ac.jp}
\orcid{0000-0001-6473-8221}
\affiliation{%
 \institution{University of Tsukuba}
 \city{Tsukuba}
 \country{Japan}
}

\author{Lei Chen}
\authornote{Also with Hong Kong University of Science and Technology.}
\email{leichen@cse.ust.hk}
\orcid{0000-0002-8257-5806}
\affiliation{%
 \institution{Hong Kong University of Science and Technology (Guangzhou)}
 \city{Guangzhou}
 \country{China}
}

\author{Xiaofang Zhou}
\email{zxf@cse.ust.hk}
\orcid{0000-0001-6343-1455}
\affiliation{%
 \institution{Hong Kong University of Science and Technology}
 \city{Hong Kong SAR}
 \country{China}
}

\renewcommand{\shortauthors}{Jialiang Wang et al.}

\begin{abstract}
  High-level automation is increasingly critical in AI, driven by rapid advances in large language models (LLMs) and AI agents.
  However, LLMs, despite their general reasoning power, struggle significantly in specialized, data-sensitive tasks such as designing Graph Neural Networks (GNNs).
  This difficulty arises from (1) the \textit{inherent} knowledge gaps in modeling the intricate, varying relationships between graph properties and suitable architectures and (2) the \textit{external} noise from misleading descriptive inputs, often resulting in generic or even misleading model suggestions.
  Achieving \textit{proficiency} in designing data-aware models---defined as the meta-level capability to systematically accumulate, interpret, and apply data-specific design knowledge---remains challenging for existing automated approaches, due to their inefficient construction and application of meta-knowledge.
  To achieve meta-level \textit{proficiency}, we propose DesiGNN, a knowledge-centered framework that systematically converts past model design experience into structured, fine-grained knowledge priors well-suited for meta-learning with LLMs.
  To account for the \textit{inherent} variability and \textit{external} noise, DesiGNN aligns empirical property filtering from extensive benchmarks with adaptive elicitation of literature insights via LLMs.
  By constructing a solid meta-knowledge between unseen graph understanding and known effective architecture patterns, DesiGNN can deliver top-5.77\% initial model proposals for unseen datasets within seconds and achieve consistently superior performance with minimal search cost compared to baselines.
\end{abstract}




\maketitle

\section{Introduction}
\label{sec:introduciton}

\begin{figure*}[t!]
    \centering
    \begin{subfigure}[t]{0.86\textwidth} 
        \centering
        \includegraphics[width=\textwidth]{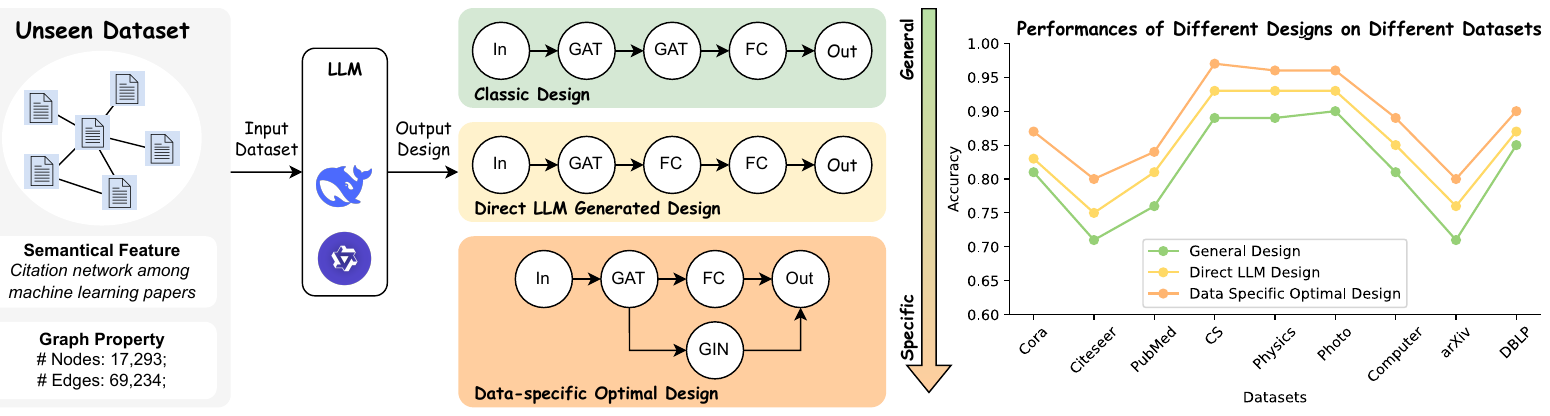}
        \caption{\textbf{Inherent Knowledge Gap:} LLMs propose overly generic GNNs when provided with purely semantic and property descriptions, unable to induce data-specific configurations.}
        \label{subfig:design-issue}
    \end{subfigure}
    \begin{subfigure}[t]{0.86\textwidth}  
        \centering
        \includegraphics[width=\textwidth]{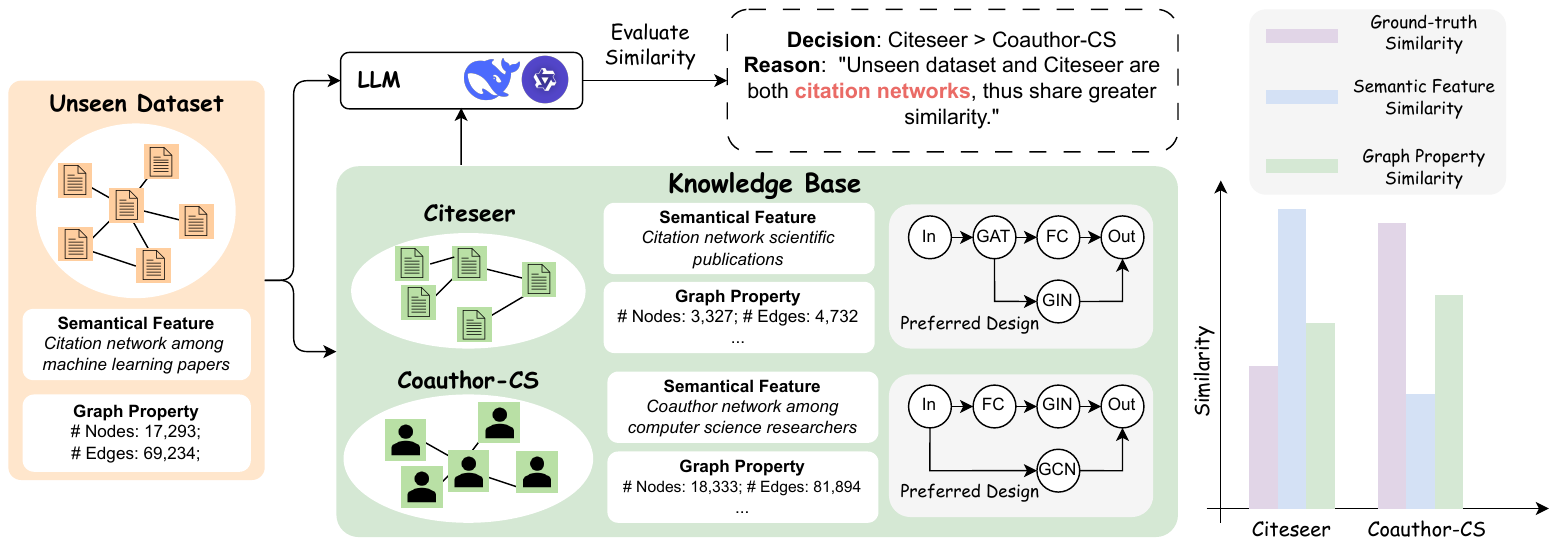}
        \caption{\textbf{External Noise}: LLMs misinterpret task similarity by weighing more on the citation networks category, causing ``artificial hallucinations'' and ineffective knowledge transfers.}
        \label{subfig:semantic-features}
    \end{subfigure}
    \vspace{-10px}
    \caption{Case studies illustrating \textit{inherent} and \textit{external} limitations of existing LLM-based methods.}
    \label{fig:overview}
    \vspace{-7px}
\end{figure*}

Automation has become a defining trend in artificial intelligence, reshaping how complex tasks are executed at scale without human intervention or the requirement of deep expertise.
From code generation~\cite{cheng2023gpt,choi2023aliro,zhang2023automl} and agentic workflow orchestration~\cite{zhang2024aflow,liu2023dynamic,zhuge2024gptswarm,xu2025robustflow,ouyang2025code2mcp} to the multi-task generalization of foundation models~\cite{liu2025graph}, this paradigm shift is transforming how expertise is embedded and operationalized in AI systems.
Among various frontiers, the automated design of machine learning models stands as one of the most fundamental and impactful challenges in this automation landscape.
In the past, automated machine learning (AutoML)~\cite{wang2022automated,wang2022profiling,wei2021pooling,zhou2022auto} has over-emphasized model search specific to independent data and tasks from scratch, which does not conform to the cross-domain and cross-task automation trends.
The recent breakthroughs in large language models (LLMs)~\cite{devlin2018bert,brown2020language,touvron2023llama,yue2025don} and AI agents~\cite{krishnan2025ai,zhu2025survey,yue2025autonomous} with strong academic backgrounds and contextual reasoning have further fueled expectations---\textit{could they autonomously generate tailored models for unseen tasks, thereby democratizing machine learning?}

Inspired by LLMs' general reasoning and generation capabilities, recent studies~\cite{tornede2023automl} have leveraged LLMs to replace human involvement beyond traditional abilities of AutoML, such as task understanding~\cite{wang2023graph,dong2023heterogeneous} and problem formulation~\cite{wei2023unleashing}.
Despite democratizing the complex process of model design, realizing highly data-aware automated model design, especially in data-sensitive domains like graph learning, remains elusive for LLMs.
Unlike tasks in vision or language domains, where LLMs excel in proposing generic solutions that could dominate all tasks and data properties, designing Graph Neural Networks (GNNs)~\cite{kipf2016semi,zhang2018link,xu2018powerful,defferrard2016convolutional,hamilton2017inductive,velivckovic2017graph} for specific graph properties and graph learning tasks requires a nuanced understanding of how data topology influences model architecture~\cite{you2020design,wang2023message}. 
For example, the homophily nature behind GCN's aggregation has catastrophic effects when propagating information over heterophily graphs, where two nodes with different labels are more likely to be linked.
Unfortunately, current LLM-based methods often rely entirely on LLMs' \textit{inherent} knowledge and reasoning, lacking fine-grained design knowledge of the relationships between graph-specific properties and effective GNN configurations, and are also sensitive to \textit{external} input noise.
As observed in Figure~\ref{subfig:design-issue}, the LLM agents adopted in recent methods~\cite{wang2023graph,dong2023heterogeneous,wei2023unleashing} tend to suggest architectures that are generally well-performing yet overly generic, \textit{inherently} falling short in tailoring designs to unseen graph structures.
Worse still, as shown in Figure~\ref{subfig:semantic-features}, they are vulnerable to ``artificial hallucinations''---being easily misled by \textit{external}, irrelevant user descriptions (e.g., being a citation network) or blindly enumerated data statistics that do not accurately reflect the data-aware challenges in model design, resulting in misleading knowledge association and ineffective designs.

Achieving \textit{proficiency} in designing data-aware models thus demands far more than general LLM-based model design methods, which call for \textit{a meta-level capability to systematically accumulate, interpret, and apply fine-grained, data-specific design knowledge}.
Humans, akin to training a new Ph.D. student, face substantial hurdles in accumulating this complex knowledge, often undergoing a lengthy learning process to design tailored models \textit{proficiently}.
Meta-based AutoML methods represent initial efforts toward this goal computationally, seeking transferable insights across tasks via learned task embeddings~\cite{you2020design,achille2019task2vec,le2022fisher} or surrogate predictors~\cite{wen2020neural,white2021powerful}.
These methods, however, often rely on simple, predefined meta-features that may fail to generalize robustly to complex data diversity, thus bending the current automation trend.
Extending this concept to graphs, AutoTransfer~\cite{autotransfer} has made the first promising strides by creating task-model repositories and embedding-based knowledge transfers to improve search efficiency.
Nonetheless, it remains limited by predefined task embeddings and the rigid aggregation of prior tasks into random search, insufficiently capturing the fine-grained, graph-aware meta-knowledge documented in the vast graph learning literature~\cite{you2020design,wang2023message}.

To address these limitations, we propose DesiGNN, a knowledge-centered and proficiency-oriented framework for the automated design of data-aware GNNs for unseen tasks.
DesiGNN systematically converts past model design experiences into structured, fine-grained knowledge priors well suited to meta-learning with LLMs, thereby novelly bridging the long-standing gap between unseen graph understanding and known effective architecture patterns in graph application research.
To address the observed \textit{external} noise from descriptive inputs and the \textit{inherent} variability in graph properties, DesiGNN constructs an empirically validated meta-level understanding of graph-GNN-performance from benchmarks and adaptively aligns key properties with graph-specific insights elicited from the graph learning literature, using LLMs as knowledge aligners to enhance the constructed meta-knowledge.
This hybridization of empirical knowledge and LLM-guided reasoning empowers DesiGNN to deliver immediate, data-aware model proposals without any training, efficiently refine these architectures based on accumulated insights, and ultimately deliver superior performance with minimal computational overhead.
Our contributions are as follows:
\begin{itemize}[leftmargin=*]
    \item We empirically identify two key limitations in current LLM-based GNN design approaches: the \textit{external} noise from descriptive inputs and the \textit{inherent} gap in fine-grained design knowledge. Motivated by these insights, we propose a knowledge-centered framework that enhances meta-level \textit{proficiency} in designing tailored, data-aware GNNs for unseen tasks by systematically accumulating, interpreting, and applying fine-grained meta-knowledge.
    \item To account for the \textit{inherent} variability of graph properties on GNN performance, DesiGNN aligns empirical property filtering from extensive benchmarks with adaptive elicitation of literature insights via LLMs, building a solid meta-knowledge between graph topology and high-performing GNN architectures. This enables DesiGNN to \textit{proficiently} deliver immediate, data-aware GNN recommendations without any training on the unseen data.
    \item Extensive experiments across 3 out-of-distribution datasets and 8 benchmarks demonstrate that DesiGNN delivers initial GNN proposals ranking in the top 5.77\% of all possible architectures within seconds. Moreover, its knowledge-driven refinement strategy rapidly optimizes these initial suggestions, consistently outperforming automated methods in both performance and efficiency.
\end{itemize}

\section{Related Work}
\label{sec:related_work}

\begin{figure*}[!t]
  \includegraphics[width=0.95\linewidth]{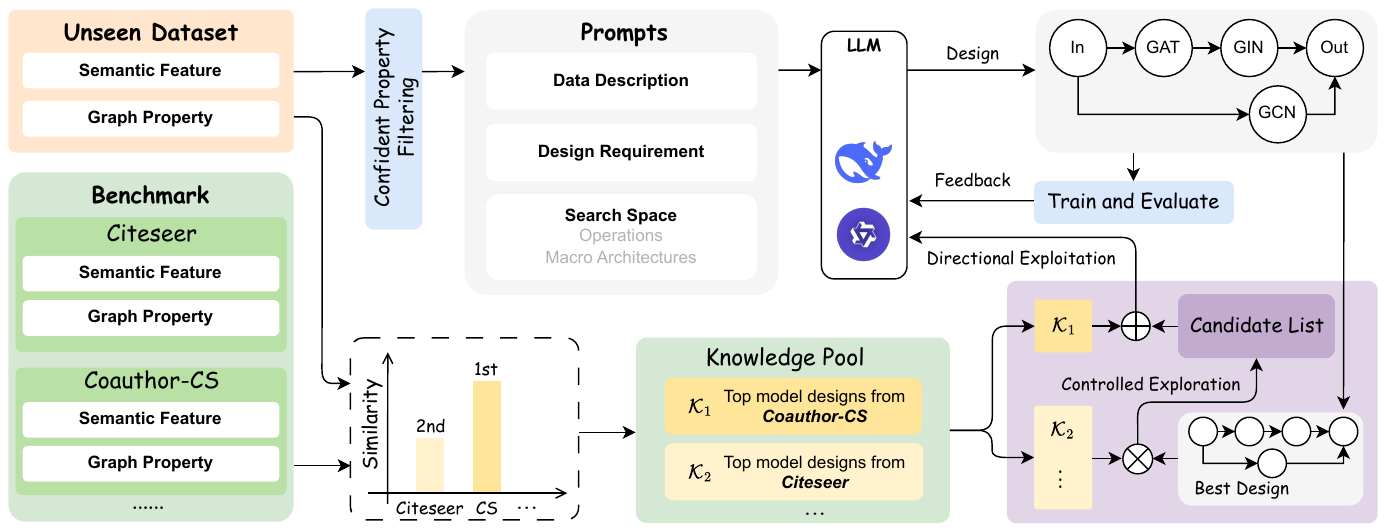}
  \vspace{-5px}
  \caption{The illustration of our knowledge-centered DesiGNN pipeline for designing GNNs. The orange module is Graph understanding, the green module is Knowledge Retrieval, and the purple module is Model Suggestion and Refinement.}
  \label{fig:pipeline}
  \vspace{-5px}
\end{figure*}

\subsection{Automated GNNs}
\label{ssec:autognn}
Automated GNNs (AutoGNNs)~\citep{wang2022automated,gao2021graph,zhou2022auto} aim to automatically find an optimal GNN architecture $\theta$ for an unseen graph $G^u$ using a controller $\pi$ parameterized by $\alpha$, where the architecture $\theta$ with network weight $\omega$ can achieve best performance $\mathcal{H}(\theta, \omega; G^u)$ on $G^u$.
In AutoGNNs, the search spaces are categorized into intra-layer~\citep{gao2019graphnas} and inter-layer~\citep{huan2021search} designs, with strategies including reinforcement learning~\citep{gao2019graphnas}, evolutionary algorithms~\citep{shi2022genetic}, differentiable search~\citep{wang2021autogel}, and search-based model fine-tuning~\citep{zhili2024search}.
Similar strategies exist for the knowledge graph~\citep{di2023message,shimin2021efficient,di2021searching,iyer2022dual}.
However, these approaches lack \textit{proficiency} and demand substantial computational resources~\citep{oloulade2021graph}, as they optimize configurations $\theta$ without leveraging prior knowledge and require extensive sampling from $\pi_{\alpha}(\theta)$.
Recently, meta-learning~\citep{finn2017model,chen2021meta} methods have sought to overcome these limitations by transferring insights from past tasks through learned task embeddings~\citep{autotransfer,achille2019task2vec} or surrogate predictors~\citep{wen2020neural,white2021powerful,li2023meta}.
While these methods mark an initial step toward knowledge-driven AutoGNNs, they rely on coarse, predefined meta-features that are often incapable of capturing generalizable relationships between diverse data properties and effective model configurations.

\textbf{NAS Benchmark for Designing GNNs.}
NAS-Bench-Graph~\citep{qin2022bench} offers a standard benchmark for GNN architecture search, comprising nine datasets and 26,206 unique architectures with associated performance records.
Similarly, many other benchmarks have been created by pioneers across different domains~\citep{chitty2023neural}.
While these benchmarks could provide valuable insights into model-performance relationships, existing works underutilize their potential for knowledge-driven design because they lack a data dimension, which is the foundation of knowledge transfer.
Therefore, only a few methods~\citep{liustructuring,you2020design,autotransfer} attempted to design feasible models for unseen tasks using such model-performance knowledge.
Notably, they rely primarily on semi-supervised or supervised techniques to fill in missing data associations, i.e., require direct training on unseen data.

\subsection{LLMs for Designing GNNs and Limitations}
\label{ssec:llm_design_gnn}
Inspired by LLMs' ability to solve optimization tasks~\citep {yang2023large}, recent studies explore their use to automate model design~\citep{zheng2023can,jawahar2023llm,zhang2023automl,yu2023gpt,zhang2023mlcopilot,guo2024ds}, focusing on managing task formulation, architecture generation, and optimization.
Auto2Graph~\citep{wei2023unleashing} democratizes the use of AutoGNNs by translating descriptive user inputs (e.g., Cora~\citep{sen2008collective} is a citation network) into configuration specifications. GPT4GNAS~\citep{wang2023graph} and GHGNAS~\citep{dong2023heterogeneous} iteratively refine model designs through crafted prompts, focusing on different graph geneity.
Despite democratizing the usage, they suffer from:
(1) Over-reliance on \textit{external} descriptive user inputs that result in poor alignment between graph properties and architecture designs;
(2) Raw LLMs \textit{inherently} lack deep expertise in fine-grained graph-GNN-performance relationships, limiting their ability to suggest data-aware models;
(3) Refining model designs solely based on feedback from testing models, lacking actionable, knowledge-driven search strategies.

\section{Proficient Graph Neural Network Design}
\label{sec:methodology}
As discussed in Section~\ref{sec:introduciton}, current LLM-based methods lack \textit{proficiency} in designing data-aware GNNs, with \textit{inherent} knowledge gaps and \textit{external} noise as major barriers.
To formalize these challenges, we adopt a Bayesian perspective to represent the automated GNNs design as solving the posterior belief over candidate architecture $\theta_i$ when facing an unseen graph $G^u$:
\begin{equation}
\label{eq:bayesian}
\mathbb{P}(\theta_i \vert G^u) = \frac{\mathbb{P}(G^u \vert \theta_i)\,\mathbb{P}(\theta_i)}{\mathbb{P}(G^u)},
\end{equation}
where $\mathbb{P}(\theta_i)$ is derived from a knowledge pool $\mathcal{K}$ accumulated from prior tasks $\{G^i\}$.
However, \textit{it remains a non-trivial problem to solve pseudo-likelihood $\mathbb{P}(G^u \vert \theta_i)$ without formally testing the model on unseen data}---a common \textit{proficiency} bottleneck in existing AutoML that leads to exhaustive search.
This \textit{proficiency} denotes the capability to systematically accumulate, interpret, and apply fine-grained, data-aware meta-knowledge relating graph properties to effective GNN architectures.
In our formulation, such capability can be conceptually represented as solving for the posterior belief $\mathbb{P}(\theta_i \vert G^u)$, with minimal or even no testing of candidate models $\{\theta_i\}$ on $G^u$.

To enhance such meta-level \textit{proficiency} without directly testing the unseen data, DesiGNN explicitly integrates empirically validated graph-property knowledge with adaptive LLM alignment to construct a trustworthy meta-level understanding $\mathcal{M}$ between graphs $\mathbf{G}$, GNNs $\mathbf{\Theta}$, and performances $\mathbf{P}$: 
$\mathcal{M}: (\mathbf{G}, \mathbf{\Theta}) \rightsquigarrow \mathbf{P}$ 
based on crucial insights from graph literature~\citep{you2020design,wang2023message}.
Figure~\ref{fig:pipeline} illustrates our Bayesian-inspired pipeline: Graph Understanding, Knowledge Retrieval, and Model Suggestion and Refinement, each component systematically addressing \textit{external} noise in descriptive data inputs and \textit{inherent} meta-knowledge gaps, and how to strategically apply the constructed meta-knowledge in designing data-aware models.

\begin{figure*}[!t]
  \centering
  \vspace{-5px}
  \begin{subfigure}[b]{0.49\textwidth}
      \centering
      \includegraphics[width=\linewidth]{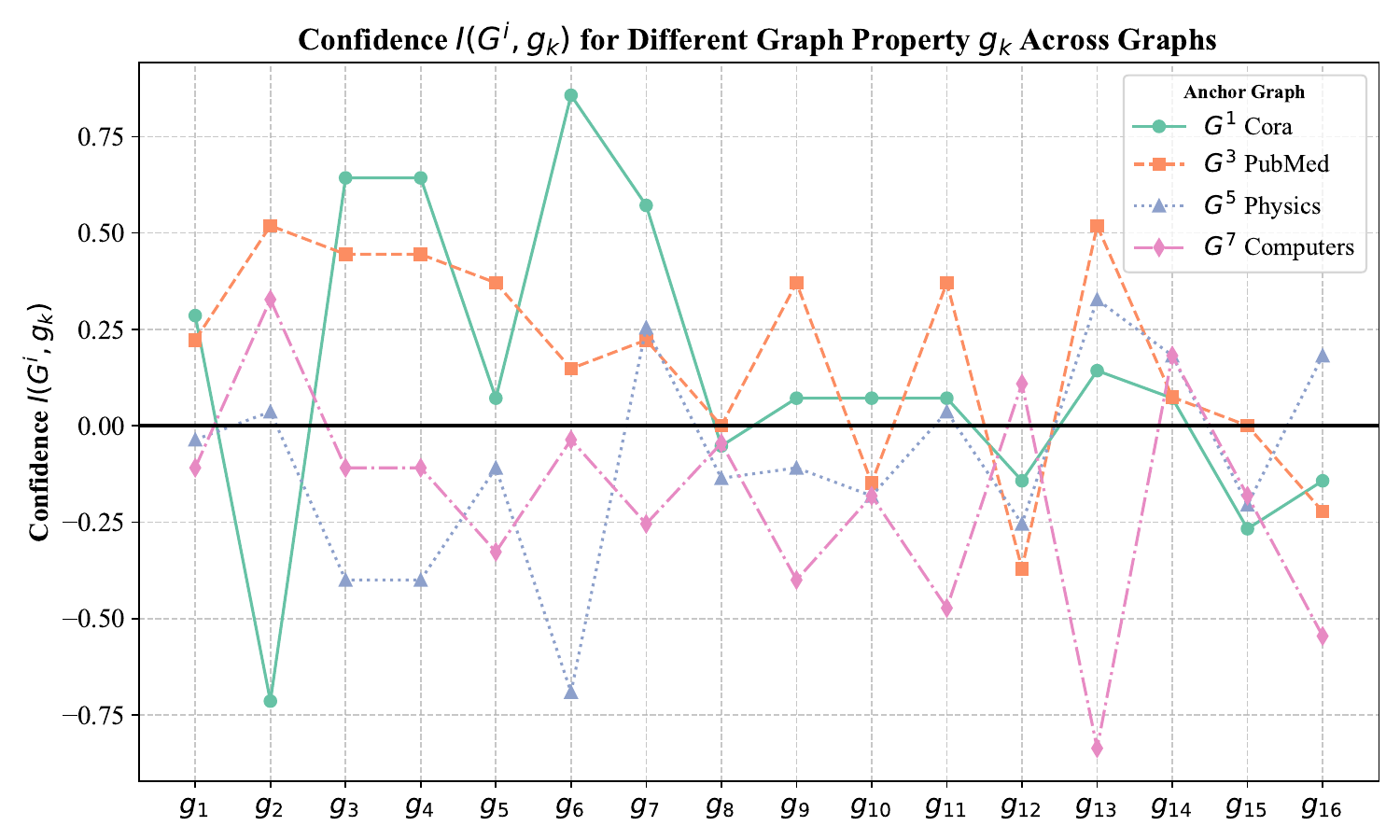}
      \caption{Property \textit{confidence} varies across graphs.}
      \label{fig:methodology_correlations}
  \end{subfigure}
  \hfill
  \begin{subfigure}[b]{0.49\textwidth}
      \centering
      \includegraphics[width=\linewidth]{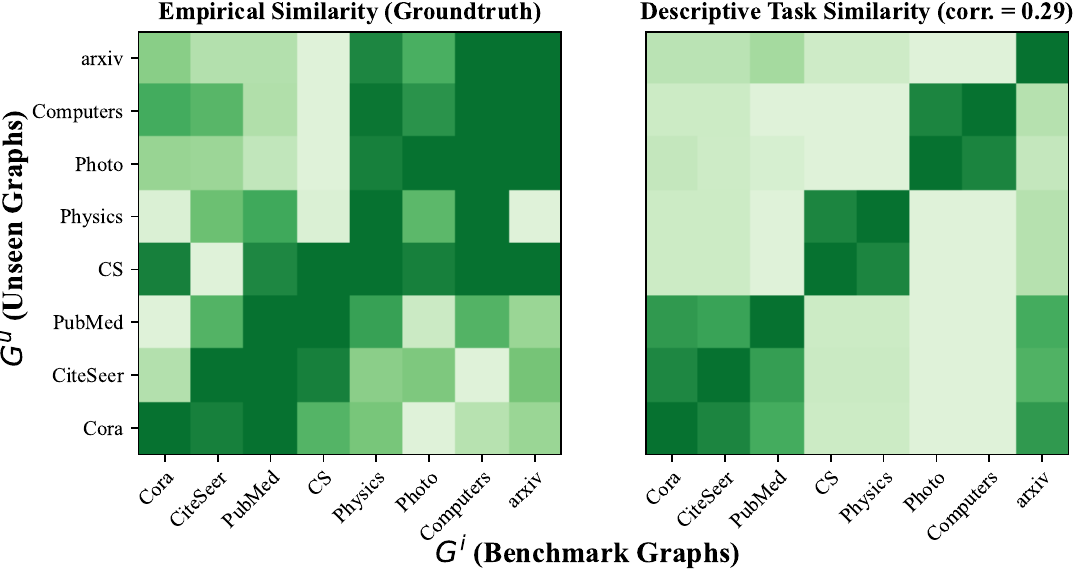}
      \caption{Descriptive inputs induce invalid task associations that deviate from empirical truth. DesiGNN exhibits a better correlation of 0.52.}
      \label{fig:descriptive}
  \end{subfigure}
  \vspace{-5px}
  \caption{Empirical studies of two barriers: \textit{inherent} property variability and \textit{external} noise.}
  \label{fig:confidence_plots}
  \vspace{-5px}
\end{figure*}

\subsection{Connect Graph Topology to GNN Design}
\label{ssec:graph_understanding}
\textit{Performance-oriented} modeling of discrete graph data is essential to quickly capture data priors from massive benchmarks to construct trustworthy meta-knowledge, a key challenge for LLMs due to their inability to naturally interpret graph-structured data~\citep{guo2023gpt4graph}.
As discussed in Section~\ref{ssec:llm_design_gnn}, existing LLM-based methods rely heavily on user-provided task descriptions, which fail to capture key graph topologies (e.g., heterophily) that often motivate specific GNN designs in the vast graph learning literature~\citep{you2020design,wang2023message}.
Our studies in Figures~\ref{subfig:semantic-features}, ~\ref{fig:descriptive}, and Section~\ref{sec:experiment_EK} demonstrate that semantic or purely descriptive inputs (e.g., a citation graph) frequently mislead LLMs, leading to ``artificial hallucination'' and ineffective solutions.
Such descriptive inputs introduce significant \textit{external} noise, undermining the reliability of generic LLM suggestions when connecting graphs to model performance.
Thus, it is necessary to establish an empirical foundation that captures fine-grained graph-GNN-performance mappings, enabling the construction of trustworthy meta-knowledge for data-aware model design.


To empirically study the missing meta-level understanding about graph-GNN-performance, we start with 16 graph properties $\{g_k\}_{k=1}^{16}$ from the literature: clustering, betweenness, density, degree centrality, closeness, degree, edge count, graph diameter, shortest path, assortativity, eigenvector, dimensionality, node count, diversity, connected component, and label homophily.
We infer graph similarity from two aspects: 1) the statistical distance ranking $\mathcal{SR}(i,k)$ based on graph properties $g_k$, and 2) the empirical performance ranking $\mathcal{ER}(i)$ of transferred top-performing GNNs from other graphs $\{G^j \mid j \neq i\}$ to anchor graph $G^i$.
To assess how reliably specific graph properties predict successful GNN design knowledge transfers, we introduce the concept of graph property \textit{confidence} for $g_k$ on $G^i$ through Kendall's $\tau$ Coefficient~\citep{kendall1938new}:
\begin{equation}
    I(G^i, g_k) := \text{KendallCorr}\big(\mathcal{SR}(i,k), \mathcal{ER}(i)\big),
\end{equation}
where $I(G^i, g_k)$ quantifies the correlation between the topological distance measured by $g_k$ and the empirical performance differences in transferring different top-performing GNNs from $\{G^j \mid j \neq i\}$ to $G^i$.
A higher $I(G^i, g_k)$ indicates that $g_k$ is a stronger indicator of effective model design transfers, thus playing a key role in meta-learning with LLMs to design data-aware models for unseen tasks.

\noindent
\textbf{Variability in Property Confidence.}
To validate the reliability of graph properties for GNN performance, we assess 16 \textit{confidence} scores $\{I(G^i, g_k)\}_{k=1}^{16}$ for each anchor graph $G^i$ against the other 8 graphs in the benchmark~\citep{qin2022bench}.
Figure~\ref{fig:methodology_correlations} shows that \textit{confidence} $I(G^i, g_k)$ fluctuates widely across diverse graphs. 
e.g., while $I(\text{Physics}, g_{16}\text{=homophily})$ is positive, the same property exhibits negative correlations for other graphs.
\textit{This inherent variability highlights that specific graph properties are strong indicators of transferability for certain graphs but unreliable for others.}
Notably, this directly exposes why generic LLM suggestions fail to capture fine-grained insights---LLMs cannot implicitly discern when particular properties matter without explicit, empirically validated priors.

\begin{figure}[!t]
  \centering
  \vspace{-5px}
  \centering
  \includegraphics[width=0.95\linewidth]{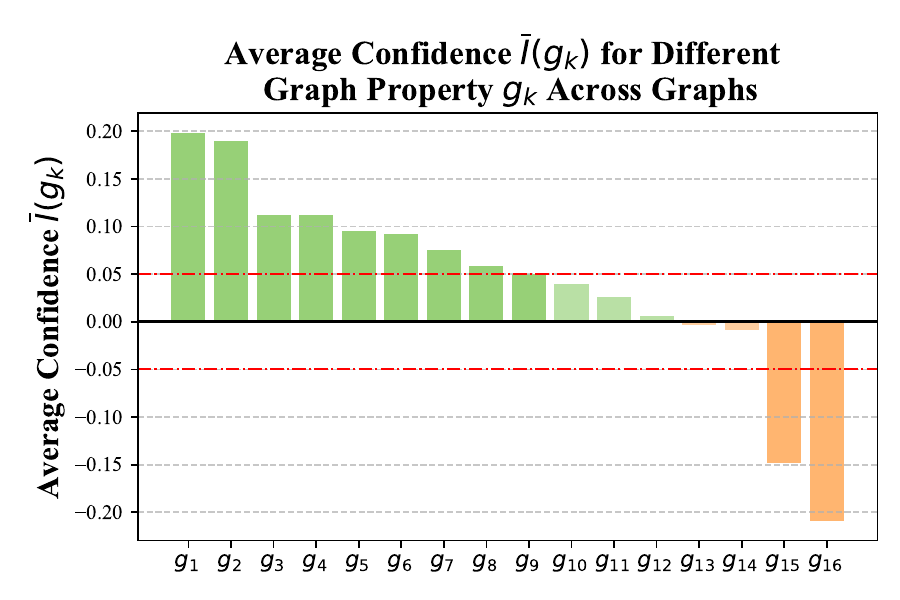}
  \vspace{-10px}
  \caption{Average \textit{confidence} of graph properties.}
  \label{fig:barchart}
  \vspace{-10px}
\end{figure}

\noindent
\textbf{Empirical Filtering as Trustworthy Foundation.}
Recognizing those empirical insights on \textit{external} noise and \textit{inherent} property variability, we first use empirical filtering to preliminarily identify graph properties that are reliably influential by averaging their predictive \textit{confidences} $I(G^i, g_k)$ across benchmark graphs as
$
\bar{I}(g_k) = \frac{1}{n} \sum_{i=1}^{n} I(G^i, g_k).
$
This process ensures that our property set built from the literature is also empirically grounded, reducing susceptibility to \textit{external} noise and \textit{inherent} property variability.
Then, based on the empirical study in Figure~\ref{fig:barchart}, we develop a filter $\mathcal{F}(G^i, \bar{I})$ to correlate the discrete graph modeling with transferred model performance, selecting the Top-$N_f$ properties with the highest $\bar{I}(g_k)$.
This leads to the alternative formulation of Equation~\ref{eq:bayesian}:
\begin{equation}
  \mathbb{P}(\theta_i \vert \mathcal{F}(G^u)) = \frac{\mathbb{P}(\mathcal{F}(G^u) \vert \theta_i)\,\mathbb{P}(\theta_i)}{\mathbb{P}(\mathcal{F}(G^u))}.
\end{equation}
This mechanism mitigates the typical LLM failure of ``artificial hallucinations''---often triggered by unprofessional graph modeling---at negligible offline computational cost.
To accumulate new knowledge, we design a self-evaluation bank $BE$ that stores measured graph properties and similarity rankings for new graphs on the fly, enabling $\mathcal{F}$ to scalably evolve over time by integrating new empirical evidence.
Grounded in empirical transferability, this mechanism serves as a trustworthy foundation for capturing graph-to-graph ($G^i$ to $G^j$) meta-priors that underpin the correlations between graphs and high-performing GNN models, yet it still \textit{calls for an effective property variability handler}.

\subsection{Knowledge Retrieval Module}
\label{sec:methodology_KR}
\textit{Proficiency} in GNN design largely depends on capturing meta-knowledge in the form of the posterior belief for unseen tasks.
However, even after empirically filtering and establishing a general correlation between graph properties and model performance, solving pseudo-likelihood $\mathbb{P}(\mathcal{F}(G^u) \vert \theta_i)$ remains a resource-intensive accumulation of testing various $\theta_i$ on $G^u$.
To overcome it, we novelly propose to leverage the known likelihood $\mathbb{P}(\mathcal{F}(G^i) \vert \theta_i)$ on benchmark datasets to form a practical proxy for $\mathbb{P}(\mathcal{F}(G^u) \vert \theta_i)$ on the unseen.
Eliciting such fine-grained model design knowledge paves the way for an efficient approximation of the posterior $\mathbb{P}(\theta_i \vert \mathcal{F}(G^u))$ with accumulated prior knowledge.
Finding such a proxy for the unseen task among the massive benchmark knowledge requires an unsupervised task similarity formulation that properly assesses the likelihood of successful model design knowledge transfer from $G^i$ to $G^u$, which \textit{has to reflect the performance outcomes $\mathcal{H}(\cdot, \cdot; G^u)$}.

\begin{definition}[Task Similarity for Model Design Knowledge Transfer]
\label{def:task_similarity}
Given an unseen graph $G^u$ and two benchmark graphs $G^i$ and $G^j$, each associated with $N_m$ top-performing models $\{\theta^*_m\}_{m=1}^{N_m} \in \Theta$, an effective task similarity $\mathcal{S}(\cdot, \cdot)$ for transferring model design knowledge satisfies:
\begin{equation}
  \begin{aligned}
    &\mathcal{S}(G^u, G^i) \geq \mathcal{S}(G^u, G^j) \\
    \Rightarrow \quad &\mathbb{P} \Bigl[\mathcal{H}(\{\theta^*_{im}\}, \cdot, G^u) \geq \mathcal{H}(\{\theta^*_{jm}\}, \cdot, G^u)\Bigr] \geq \delta.
  \end{aligned}
\end{equation}
where $\mathcal{H}(\{\theta^*_{im}\}, \cdot, G^u)$ denotes the performance of the transferred top-performing pattern $\{\theta^*_{im}\}$ from benchmark $G^i$ on the unseen $G^u$, and $\delta$ is a confidence threshold.
This ensures that higher task similarity between $G^u$ and $G^i$ implies a larger likelihood of high-performing transfer from $G^i$ to $G^u$.
\end{definition}

To design \textit{unsupervised} task similarity between $G^u$ and $G^i$, we first apply the filter $\mathcal{F}(\cdot)$ on $G^u$ and $G^i$ to establish an empirical foundation based on \textit{performance-oriented confidence} $\bar{I}(g_k)$.
However, as discussed in Section~\ref{ssec:graph_understanding}, graph properties \textit{inherently} exhibit varying levels of \textit{confidence} across graphs, making it challenging to transfer knowledge reliably with standalone statistical properties.
To account for \textit{inherent} variability, we further introduce the adaptive importance $w^u_k$ for each property $g_k$ specifically tailored to the unseen graph $G^u$ and define our unsupervised task similarity as: 
\begin{equation}
\label{eq:task_similarity}
  \mathcal{S}(G^u, G^i) = \frac{1}{N_f} \sum_{k=1}^{N_f} w^u_k \cdot \frac{\bar{I}(g_k)}{1 + \hat{d_k^{ui}}},
\end{equation}
where $N_f$ is the number of influential graph properties selected via empirical filtering, and $\bar{I}(g_k)/(1 + \hat{d_k^{ui}})$ measures the \textit{confident} performance similarity of property $g_k$ between graphs $G^u$ and $G^i$.

\textbf{Adaptive Knowledge Elicitation and Alignment.}
While $\bar{I}(g_k)$ is grounded in the observations from benchmarks, their adaptation to unseen tasks requires finding the adaptive importance of property $w^u_k$, which cannot be observed without testing on $G^u$.
Notably, such adaptive importance of property for a given graph commonly motivates specific GNN designs in the graph learning literature~\citep{you2020design,wang2023message}, e.g., GraphSAGE~\citep{hamilton2017inductive} handles high-degree graphs.
Therefore, we novelly propose to leverage LLMs---pre-trained on a vast corpus of graph learning literature---as meta-controllers to adaptively elicit $w^u_k$ for $G^u$ by aligning task descriptions, filtered graph properties, and high-performing architecture patterns in textual space, followed by contextual reasoning~\citep{min2022rethinking}.
This process ensures that the data-specific weights $w^u_k$ reflect interpretable insights from the literature on graph properties, GNNs, and performance, aligning with general empirical evidence to construct trustworthy meta-knowledge that empowers LLMs to quickly design effective, data-aware models for unseen tasks.

\noindent
\textbf{Knowledge Pool Construction.}
\label{sssec:knowledge_pool}
Using the computed task similarity set $\mathcal{S}^u$ for $G^u$, we rank the benchmark graphs and select the Top-$N_s$ most similar sources. The knowledge pool $\mathcal{K}$ for $G^u$ is then constructed by aggregating the top-performing models from them:
\begin{equation}
  \mathcal{K} = \bigcup_{G^i \in \text{Top-}{N_s}(\mathcal{S}^u)} \{(G^{i}, \{\theta_{im}^*\}_{m=1}^{N_m})\},
\end{equation}
The top-performing models $\{\theta_i^*\}$ act as a truncated prior $\mathbb{P}(\theta_i^*)$, summarizing accumulated design knowledge from benchmark $G^i$, while task similarity $\mathcal{S}(G^u, G^i)$ serves as a gating module for solving pseudo-likelihood $\mathbb{P}(\mathcal{F}(G^u) \vert \theta_i^*)$, indicating the effectiveness of $\theta_i^*$ on unseen graph $G^u$.

\begin{table*}[t]
  \caption{
  Initial performance comparison (accuracy) of GNNs designed by different methods. $N_s=3$ for \texttt{Desi-Init}. Best overall results are in \textbf{bold}, and best among each category are \underline{underlined}.
  }
  \vspace{-8px}
  \label{tab:initial_benchmark}
  \centering
  \setlength\tabcolsep{1.7pt}
  \resizebox{\textwidth}{!}{
  \begin{tabular}{lcccccccccccc}
  \toprule
  \textbf{Type} & \textbf{Model} & \textbf{Cora} & \textbf{Cit.} & \textbf{Pub.} & \textbf{CS} & \textbf{Phy.} & \textbf{Pho.} & \textbf{Com.} & \textbf{arX.} & \textbf{DBL.} & \textbf{Fli.} & \textbf{Act.} \\
  \midrule
  
  \multirow{7}{*}{\makecell{Manual\\GNNs}}
  & GCN & \underline{80.97(0.39)} & 69.90(1.26) & \underline{77.46(0.61)} & 88.65(0.57) & 90.85(1.20) & 89.44(0.48) & \underline{83.16(0.55)} & 71.08(0.16) & 84.25(0.25) & \underline{54.27(0.14)} & 24.78(1.79) \\
  & GAT & 80.83(0.47) & \underline{70.70(0.71)} & 75.93(0.26) & 88.72(0.73) & 89.47(1.14) & \underline{89.93(1.75)} & 81.35(1.26) & \underline{71.24(0.10)} & 84.98(0.15) & 51.85(0.26) & 26.91(1.09) \\
  & SAGE & 79.47(0.31) & 66.13(0.90) & 75.50(1.14) & 87.81(0.18) & \underline{91.43(0.29)} & 88.29(1.03) & 81.46(0.73) & 70.78(0.17) & \underline{85.41(0.06)} & 52.84(0.19) & 30.13(0.70) \\
  & GIN & 79.77(0.38) & 63.30(1.26) & 76.74(0.86) & 81.08(3.09) & 86.67(0.86) & 87.37(1.01) & 73.95(0.16) & 61.33(0.70) & 82.82(0.82) & 51.08(0.08) & 27.15(0.38) \\
  & ChebNet & 79.40(0.57) & 67.03(1.02) & 75.13(0.49) & 89.50(0.36) & 89.75(0.87) & 86.65(0.77) & 79.10(2.26) & 70.87(0.10) & 84.84(0.21) & 53.75(0.16) & 30.46(0.77) \\
  & ARMA & 78.33(0.69) & 66.20(0.75) & 75.00(0.51) & \underline{89.87(0.35)} & 88.88(1.09) & 86.55(3.35) & 78.47(0.57) & 70.87(0.17) & 84.31(0.30) & 54.23(0.04) & 31.29(0.43) \\
  & k-GNN & 78.06(0.47) & 30.97(3.56) & 75.38(0.97) & 83.81(0.58) & 88.98(0.54) & 86.45(0.21) & 76.31(1.34) & 63.18(0.38) & 83.59(0.07) & 51.18(0.33) & \underline{32.74(0.68)} \\
  \midrule
  
  \multirow{3}{*}{\makecell{Auto.\\NNI}}
  & Random & 77.87(2.41) & 66.64(1.32) & 74.16(1.68) & 81.78(9.41) & 90.59(0.94) & 89.04(2.55) & 76.61(3.56) & 68.93(1.82) & 76.45(7.53) & 52.50(0.72) & 30.90(3.75) \\
  & EA & 78.23(1.04) & 66.40(2.63) & 72.88(2.11) & 87.03(2.64) & 88.07(2.41) & 87.30(1.38) & 77.56(6.42) & 68.28(2.95) & \underline{85.13(0.38)} & 53.22(0.92) & 31.69(2.95) \\
  & RL & 73.44(8.11) & 65.35(2.40) & 75.44(1.24) & 86.17(5.09) & 88.15(4.24) & 89.48(1.35) & 77.70(3.07) & 68.00(4.71) & 84.18(0.50) & 52.07(2.84) & 31.72(4.60) \\
  \midrule
  
  \multirow{2}{*}{\makecell{Auto.\\AutoGL}}
  & GNAS & 78.55(1.20) & 63.25(5.87) & 73.04(1.64) & 86.04(7.88) & 89.54(1.52) & 87.27(2.96) & 70.96(9.66) & 69.94(1.71) & 84.46(0.41) & \underline{54.67(0.54)} & 32.31(3.25) \\
  & Auto-GNN & 78.58(2.18) & 65.60(2.69) &  \underline{76.07(0.77)} & 89.06(0.42) & 89.26(1.51) & 89.34(1.75) & 77.49(3.41) & 67.62(1.72) & 84.67(0.62) & 50.97(1.21) & 30.18(4.67) \\
  \midrule

  \multirow{3}{*}{\makecell{Meta\\-based}}
  & Kendall & 67.73 & \underline{69.20} & 71.80 & 88.56 & \underline{91.56} & 88.90 & 76.85 & 71.49 & - & - & - \\
  & Overlap & 79.36 & 67.30 & 71.80 & 88.56 & 89.95 & 90.37 & 76.85 & \underline{71.68} & - & - & - \\
  & AutoTrans. & \underline{80.24(0.72)} & 68.19(1.22) & 76.04(0.86) & 89.19(0.77) & 91.19(1.04) & \underline{90.93(0.37)} & \underline{81.77(1.25)} & 71.37(0.39) & - & - & - \\
  \midrule
  
  \multirow{2}{*}{\makecell{LLM\\-based}}
  & GPT4GNAS & 78.50(0.37) & 67.46(0.76) & 73.89(0.86) &  \underline{89.26(0.38)} & 89.44(1.94) & 89.12(2.26) & 77.21(5.26) & 68.98(1.22) & 84.93(0.22) & 52.47(0.10) &  \underline{34.26(0.47)} \\
  & GHGNAS & 79.13(0.45) & 67.35(0.44) & 74.90(0.57) & 89.15(0.81) & 88.94(2.57) & 89.42(1.99) & 77.04(3.96) & 69.66(1.28) & 85.06(0.15) & 52.48(0.23) & 33.72(2.72) \\
  \midrule
  
  \multirow{2}{*}{Ours}
  & \texttt{Desi-Init} & 80.31(0.00) & 69.20(0.16) & {76.60(0.00)} & {89.64(0.08)} & {92.10(0.00)} & {91.19(0.00)} & {82.20(0.00)} & 71.50(0.00) & {85.56(0.24)} & {55.16(0.11)} & {34.41(0.48)} \\
  & \textbf{\texttt{DesiGNN}} & \textbf{81.77(0.40)} & \textbf{71.00(0.09)} & \textbf{77.57(0.29)} & \textbf{90.51(0.42)} & \textbf{92.61(0.00)} & \textbf{92.38(0.06)} & \textbf{84.08(0.66)} & \textbf{72.02(0.18)} & \textbf{85.89(0.21)} & \textbf{55.44(0.06)} & \textbf{37.57(0.62)} \\
  \bottomrule
  \end{tabular}
  }
  \vspace{-5px}
\end{table*}

\subsection{Model Suggestion and Refinement}
\label{sec:methodology_MSR}
Overall, our formulation of Knowledge Pool $\mathcal{K}$ approximates the posterior $\mathbb{P}(\theta_i^* \vert \mathcal{F}(G^u))$, prioritizing models that align with data-aware evidence.
This Bayesian-inspired process ensures solid use of constructed meta-knowledge for designing models in both one-shot and a few attempts.

\noindent
\textbf{Initial Model Suggestion.}
\label{sec:methodology_IMS}
By supplying contextual meta-knowledge of high similarities to the unseen graph $G^u$ from $\mathcal{K}$, we leverage LLM as a surrogate meta-controller to sample $N_s$ initial proposals $\{\theta_{ui}\}_{i=1}^{N_s}$ based on the approximated posterior,
where $\mathcal{K}_i$ represents the specific knowledge from the $i$-th similar benchmark graph in $\mathcal{K}$:
$
\theta_{ui} \leftarrow \mathcal{LLM}_{IMS}(\mathcal{F}(G^u, \bar{I}), \mathcal{K}_i).
$. 
Compared to existing AutoGNNs, our method \textit{avoids any model training} on unseen data, delivering promising designs by combining fine-grained literature and empirical insights for immediate model deployment.

\noindent
\textbf{Knowledge-Driven Refinement.}
\label{sec:methodology_LGNAS}
To further optimize initial model proposals, we design a structured, knowledge-driven refinement strategy inspired by expert-like exploration-exploitation trade-offs, as illustrated in Figure~\ref{fig:pipeline}.
Our local search strategy iteratively improves models using the constructed meta-knowledge of high similarities to the unseen from $\mathcal{K}$.
\textit{(0) Re-Ranking:} Initial proposals $\{\theta_{ui}\}_{i=1}^{N_s}$ are re-ranked based on their validation performance, and the best-performing proposal $\theta_{u1}$ serves as the refinement starting point.
\textit{(1) Controlled Exploration:} Configurations from other models in $\mathcal{K}$ are crossovered with $\theta_{u1}$ to generate $N_c$ new candidates.
\textit{(2) Model Promotion Mechanism:} Candidates are ranked based on their retrieved performances on the benchmark dataset of $\mathcal{K}_1$, with the most promising candidate $\theta_{u}^{t'}$ advanced for further refinement at iteration $t$.
\textit{(3) Directional Exploitation:} The LLM meta-controller $\mathcal{LLM}_{KDR}$ mutates $\theta_{u}^{t'}$ using user requirements, task descriptions, search space details, and previous training logs.
\textit{(4) Evaluation and Update:} Each refined candidate $\theta_{u}^t$ is validated and added to the optimization trajectory $\Theta^T$. The best-performing model is updated as $\theta_{u}^*$.
This refinement process balances exploration (testing diverse configurations) and exploitation (focusing on high-potential candidates), leveraging prior knowledge in the textual space to efficiently optimize model designs.

\section{Experiments}
\label{sec:experiments}

\subsection{Experimental Settings}
\label{sec:experiment_settings}

\noindent
\textbf{Task and Datasets.}
We conduct node classification studies using 8 benchmark datasets from NAS-Bench-Graph~\citep{qin2022bench}: Cora, Citeseer, PubMed~\citep{sen2008collective}, CS, Physics, Photo, Computer~\citep{shchur2018pitfalls}, and ogbn-arXiv~\citep{hu2020open}.
We include 3 extra datasets for out-of-distribution (e.g., heterophily) generalization: DBLP~\citep{bojchevski2017deep}, Flickr~\citep{zeng2019graphsaint}, and Actor~\citep{pei2020geom}.

\noindent
\textbf{Baselines.}
We compare DesiGNN with 4 categories of 17 baselines:
\textit{(1) Manually Designed GNNs:} 7 GNNs~\citep{kipf2016semi,velivckovic2017graph,hamilton2017inductive,xu2018powerful,defferrard2016convolutional,bianchi2021graph,morris2019weisfeiler} showcasing human experts' efforts in designing tailored models for specific types of graphs;
\textit{(2) Classical Automated Approaches:} 3 typical automated search strategies~\citep{li2020random,real2019regularized,zoph2016neural}, plus 2 advanced AutoGNN systems~\citep{gao2021graph,zhou2022auto}, all operating within the same search space as DesiGNN for fairness;
\textit{(3) Meta-based Approaches:} 3 meta-based, semi-supervised methods~\citep{you2020design,qin2022bench,autotransfer} that recommend models from the most similar benchmark graphs or based on meta-knowledge after testing anchor models on the unseen task;
\textit{(4) LLM-Based Approaches:} 2 recent SOTA LLM-Based methods~\citep{wang2023graph,dong2023heterogeneous} that leverage LLMs for iterative architecture search.

\noindent
\textbf{Evaluation Metrics.}
To ensure reliability, we report the average accuracy and standard deviation over 10 runs.
For automated methods, we also analyze the best-so-far accuracy after validating 1-30 model proposals to measure short-run efficiency.
To assess efficiency under varying LLM API traffic, we set the basic time unit as the time to validate a model proposal on the unseen dataset.

\noindent
\textbf{Implementation Details.}
Our code\footnote{\label{code}\textbf{Full code, data, prompts:} \url{https://github.com/jilwang84/DesiGNN}.} is implemented in PyTorch and LangChain (GPT-4).
For clarity, \texttt{Desi-Init} refers to initial model proposals, while \texttt{DesiGNN} denotes the whole pipeline, including model refinement.
We use NAS-Bench-Graph~\citep{qin2022bench} to accumulate knowledge and provide a unified search space.
To prevent data leakage, when a benchmark dataset is treated as unseen, it is completely excluded from the knowledge retrieval process and anonymized to ensure fair evaluation.
We have validated that LLMs, given only dataset descriptions, lack prior knowledge of effective architectures in the NAS-Bench-Graph search space. 

\begin{figure*}[t]
    \centering
    \includegraphics[width=0.98\linewidth]{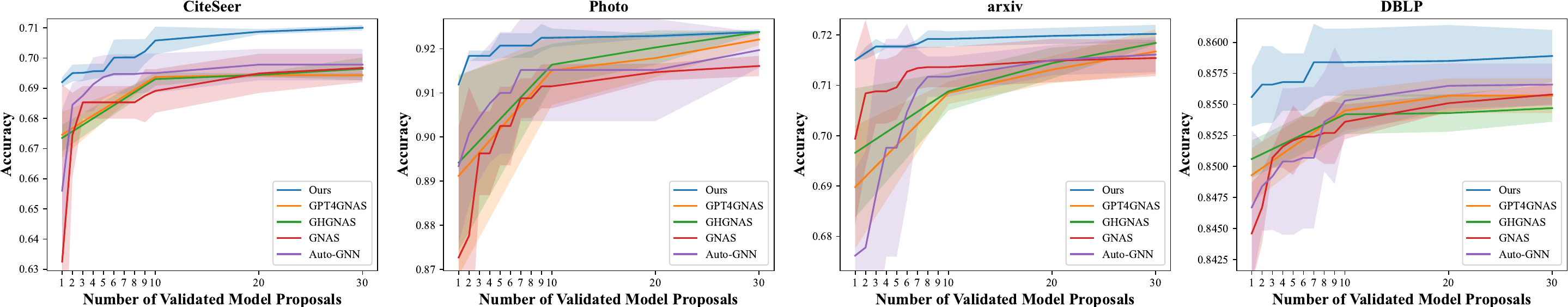}
	\vspace{-8px}
    \caption{Short-run performance of \texttt{DesiGNN} compared to stronger automated baselines.}
    \label{fig:short_run}
	\vspace{-8px}
\end{figure*}

\subsection{Main Results}
\label{exp:main_results}

\noindent
\textbf{Initial Model Suggestions.}
First, we use \texttt{Desi-Init} in Table~\ref{tab:initial_benchmark} to refer to the initial model suggestion part in Section~\ref{sec:methodology_MSR}, i.e., the first GNN suggested by LLM within seconds (avg. $0.07$ sec of DesiGNN system time cost + avg. $10.77$ sec each for LLM API communication, which may vary among users and API providers).
While human-designed GNNs like GCN and GAT excel on popular datasets (e.g., Cora, Citeseer), they fail to generalize to less common heterophily datasets such as Flickr and Actor.
In contrast, \texttt{Desi-Init} surpasses these overly specialized models, demonstrating robustness across diverse graphs.
Notably, \texttt{Desi-Init} outperforms semi-supervised meta-based baselines, validating our contributions of establishing \textit{proficient} ``graph-GNN-performance'' meta-knowledge through unsupervised task similarity assessment and graph understanding with adaptive properties.
Furthermore, compared to automated baselines, \texttt{Desi-Init} achieves superior performance across all 11 datasets, underscoring the strength of our knowledge-driven approach over conventional automated methods (which lack domain-specific insights) and LLM-based methods (which rely solely on commonplace knowledge).
Overall, DesiGNN's initial model proposals rank in the top 5.77\% of all possible architectures.

\begin{table}[t]
\centering
\caption{Computational resources (in \#proposal validations) needed by baselines to reach \texttt{DesiGNN}'s 10-validation performance. $^\ast$ contains failed cases within 100 validations.}
\vspace{-8px}
\label{tab:count_validations}
\setlength\tabcolsep{1pt}
\begin{tabular}{lcccccccccc}
\toprule
& & \multicolumn{9}{c}{\#Model Proposal Validations} \\
\cmidrule(lr){2-11}
\textbf{Model} & \textbf{Cora} & \textbf{Cit.} & \textbf{Pub.} & \textbf{CS} & \textbf{Phy.} & \textbf{Pho.} & \textbf{Com.} & \textbf{arX.} & \textbf{DBL.} & \textbf{Fli.} \\
\midrule
\textbf{GNAS} & 40.3$^\ast$ & 48.3$^\ast$ & 21.5$^\ast$ & 38.8 & $\infty$ & 66$^\ast$ & 34.8 & 55$^\ast$ & 69.4 & 77.2$^\ast$ \\
\textbf{Auto-.} & 21.8$^\ast$ & $\infty$ & 11.6 & 18.4 & 43$^\ast$ & 29.3$^\ast$ & 17$^\ast$ & 59.3$^\ast$ & 76.2$^\ast$ & $\infty$ \\
\midrule
\textbf{Rand.} & 21.7$^\ast$ & $\infty$ & 45 & 18.8 & 89$^\ast$ & 33$^\ast$ & 22 & 87$^\ast$ & 64.7 & $\infty$ \\
\textbf{EA} & 24.7$^\ast$ & $\infty$ & 48.8$^\ast$ & 40.8$^\ast$ & $\infty$ & 23$^\ast$ & 51.3$^\ast$ & 72.8$^\ast$ & $\infty$ & 46.3 \\
\textbf{RL} & 82$^\ast$ & $\infty$ & 11.8 & 14.8 & $\infty$ & 84.5$^\ast$ & 48.2 & 37$^\ast$ & 38.8 & 99$^\ast$ \\
\midrule
\textbf{G4G.} & 30-40 & $\infty$ & 40-50 & 10 & $\infty$ & 30-40 & 10-20 & 40-50 & 40-50 & $\infty$ \\
\textbf{GHG.} & 20-30 & $\infty$ & 40-50 & 10 & $\infty$ & 20-30 & 20-30 & 30-40 & $\infty$ & $\infty$ \\
\bottomrule
\end{tabular}
\vspace{-10px}
\end{table}

\begin{table*}[!t]
  \vspace{-1px}
  \caption{The main ablation study on each component in GNN Model Suggestion and Refinement. * is our complete setting.}
  \vspace{-8px}
  \label{tab:main_ablation}
  \centering
  \setlength\tabcolsep{1.7pt}
  \resizebox{\textwidth}{!}{
  \begin{tabular}{lcccccccc}
    \toprule
    & \multicolumn{8}{c}{ACC (STD) \%} \\
    \cmidrule(lr){2-9}
    \textbf{Method} & \textbf{Cora} & \textbf{Citeseer} & \textbf{PubMed} & \textbf{CS} & \textbf{Physics} & \textbf{Photo} & \textbf{Computers} & \textbf{arXiv}\\
    \toprule
    \multicolumn{9}{c}{\textbf{Initial Proposal Performance}} \\
    \midrule
    \textbf{Property Only*} & \textbf{80.31 (0.00)} & 69.20 (0.16) & \textbf{76.60 (0.00)} & \textbf{89.64 (0.08)} & \textbf{92.10 (0.00)} & \textbf{91.19 (0.00)} & \textbf{82.20(0.00)} & \textbf{71.50 (0.00)} \\
    \textbf{Descriptional Only} & \textbf{80.31 (0.00)} & \textbf{69.26 (0.00)} & 75.71 (0.00) & 89.53 (0.00) & 88.42 (0.00) & 91.17 (0.13) & 81.79 (1.21) & 71.20 (0.34) \\
    \textbf{Both} & \textbf{80.31 (0.00)} & \textbf{69.26 (0.00)} & 75.71 (0.00) & 89.55 (0.03) & 91.45 (0.78) & 91.13 (0.07) & \textbf{82.20(0.00)} & \textbf{71.50 (0.00)} \\
    \textbf{w/o Knowledge} & 79.30 (0.00) & 55.29 (0.00) & 71.56 (0.00) & 81.94 (0.00) & 91.45 (0.00) & 86.61 (0.00) & 69.32 (0.00) & 70.68 (0.00) \\
    \toprule
    \multicolumn{9}{c}{\textbf{Best Performance After 30 Validations}} \\
    \midrule
    \textbf{All*} & \textbf{81.77 (0.40)} & \textbf{71.00 (0.09)} & \textbf{77.57 (0.29)} & \textbf{90.51 (0.42)} & 92.61 (0.00) & \textbf{92.38 (0.06)} & \textbf{84.08 (0.66)} & \textbf{72.02 (0.18)} \\
    \textbf{w/o Re-rank} & \textbf{81.77 (0.40)} & 69.77 (0.31) & 77.40 (0.24) & 90.47 (0.00) & 92.61 (0.00) & \textbf{92.38 (0.06)} & 83.90 (0.14) & 71.87 (0.12) \\
    \textbf{w/o Promotion} & 81.04 (0.42) & 70.33 (0.00) & 77.00 (0.30) & 90.19 (0.47) & \textbf{92.71 (0.18)} & 92.22 (0.16) & 82.85 (0.01) & 71.99 (0.01) \\
    \textbf{w/o Exploration} & 81.61 (0.10) & 70.34 (0.64) & 77.40 (0.14) & 90.24 (0.33) & 92.61 (0.00) & 91.87 (0.18) & 83.74 (0.23) & 71.99 (0.08) \\
    \textbf{w/o Knowledge} & 80.90 (0.00) & 69.93 (0.00) & 76.72 (0.00) & 90.24	 (0.00) & 91.82 (0.00) & 92.21 (0.00) & 83.63 (0.00) & 71.71 (0.00) \\
    \bottomrule
  \end{tabular}
  }
  \vspace{-5px}
\end{table*}

\begin{figure*}[t]
    \centering
    \includegraphics[width=0.98\linewidth]{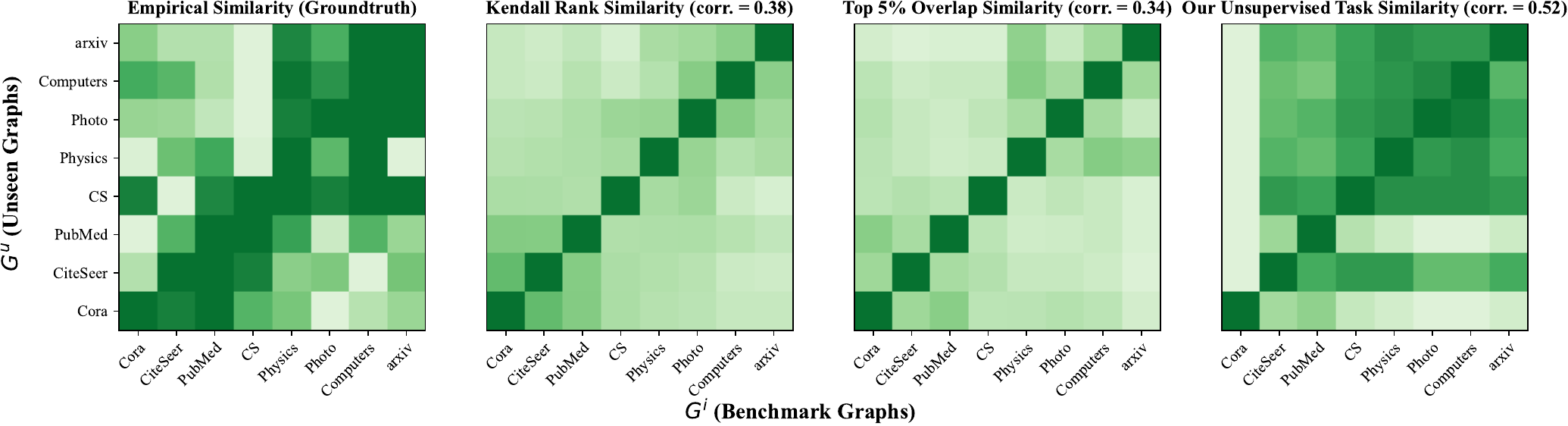}
	\vspace{-10px}
    \caption{Task similarities computed by empirical performance (target), two meta-based baselines, and ours. Kendall's $\tau$ correlation coefficient with the empirical ground truth is shown in parentheses.}
    \label{fig:heatmap}
	\vspace{-10px}
\end{figure*}

\begin{table}[t]
\centering
\caption{Hit rate $\delta^*$ across unseen datasets.}
\vspace{-8px}
\label{tab:hit_rate}
\setlength\tabcolsep{1pt}
\begin{tabular}{lccc}
\toprule
\textbf{Top-s Benchs} & $N_s$=1 & $N_s$=2 & $N_s$=3 \\
\midrule
\textbf{Kendall} & 0 & 0.25 & 0.625 \\
\textbf{Overlap} & 0 & 0.375 & 0.625 \\
\midrule
\textbf{\makecell[l]{Property Similar. (w/o LLMs)}} & \textbf{0.375} & 0.5 & 0.625 \\
\textbf{\makecell[l]{Description Only (with LLMs)}} & 0.262 & 0.5 & 0.575 \\
\textbf{\makecell[l]{Property Only (with LLMs)}} & \textbf{0.375} & \textbf{0.6} & \textbf{0.725} \\
\textbf{\makecell[l]{Both (with LLMs)}} & 0.275 & 0.5 & 0.6 \\
\bottomrule
\end{tabular}
\vspace{-7px}
\end{table}

\noindent
\textbf{Model Refinement and Short-run Efficiency.}
Building on the initial proposals, \texttt{DesiGNN} can rapidly refine models to outperform all manually designed and automated GNNs across 11 datasets, demonstrating its efficient use of meta-knowledge to exploit the high potential range.
Figure~\ref{fig:short_run} highlights \texttt{DesiGNN}'s short-run efficiency, achieving significant performance improvements with fewer model proposal validations than automated baselines.
Unlike methods that require extensive validation to identify feasible models, \texttt{DesiGNN} emulates human-like refinement strategies, reducing computational costs and accelerating optimization.
Specifically, Table~\ref{tab:count_validations} shows that other automated methods require significantly more model validations to match \texttt{DesiGNN}'s performance after just 10 validations.
These results confirm that DesiGNN effectively addresses computational inefficiencies by leveraging accumulated knowledge to enable immediate feedback and rapid optimization, delivering consistently superior performance under limited budgets.

\subsection{Effective Knowledge via Task Similarity}
\label{sec:experiment_EK}
To assess whether our unsupervised task similarity in Equation~\ref{eq:task_similarity} reflects the probability of successful knowledge transfer defined in Definition~\ref{def:task_similarity}, we quantify the \textit{hit rate}, defined as the tighter probability $\delta^*$ that the empirically most relevant benchmarks (i.e., those maximizing $\mathcal{H}(\{\theta^*_{im}\}, \cdot, G^u)$) are included in the Top-$N_s$ benchmarks retrieved by $\mathcal{S}$ ($N_s$ is the tolerance).
A higher hit rate directly supports the probabilistic claim in Definition~\ref{def:task_similarity}, where $\mathcal{S}$ correlates with the likelihood of transferring high-performing models from $G^i$ to $G^u$.
As shown in Table~\ref{tab:hit_rate}, our unsupervised task similarity based on graph properties achieves the highest hit rate across all tolerance settings, outperforming methods that rely on Kendall or Top 5\% similarities, as well as purely statistical property similarity (without LLMs to conduct adaptive knowledge elicitation and alignment).
This demonstrates that our \textit{performance-oriented} $\mathcal{S}$, derived from key graph properties and adaptive weighting by LLMs, effectively captures the meta-knowledge required for data-aware model design.
Besides, we observed a negative effect when incorporating descriptive inputs into similarity assessment, suggesting that some semantic properties (e.g., being a citation graph) are less relevant to designing the model architecture (see the case study in Figure~\ref{subfig:semantic-features}).
We further visualize the pairwise similarities between graphs in Figures~\ref{fig:descriptive} and ~\ref{fig:heatmap}, showing that our unsupervised task similarity closely aligns with empirical performance similarity (the highest Kendall's $\tau$ correlation).
This further validates the claim that $\mathcal{S}$ approximates a pseudo-likelihood $\mathbb{P}(\mathcal{F}(G^u) \vert \theta_i^*)$ well, reinforcing its role in supporting the Bayesian-inspired framework in Section~\ref{sec:methodology_KR}.


\subsection{Main Ablation Study}
\label{sec:ablation_Main}
We present the main ablation study on the Knowledge Retrieval and the GNN Model Suggestion and Refinement module in Table~\ref{tab:main_ablation}.
The results from the initial model suggestion stage underscore the importance of designing an effective Graph Understanding method, demonstrating that relying solely on filtered graph properties in dataset descriptions outperforms using descriptional inputs.
In the short run, after 30 model validations, the Re-rank mechanism significantly improves the knowledge-driven model refinement process when the original order of the Knowledge Pool $\mathcal{K}$ is incorrect.
Additionally, the Model Promotion mechanism, which simulates the strategy of human experts refining a model based on accumulated knowledge, plays a crucial role in enhancing the efficacy of model refinement.
Lastly, the Directional Exploration mechanism is empirically beneficial, as it leverages the most relevant model design knowledge and the in-context learning ability of LLMs to further refine the best candidate model promoted.

\begin{table}[!t]
  \caption{Different LLMs for model suggestion and refinement.}
  \vspace{-8px}
  \label{tab:llm_comparison_combined}
  \setlength\tabcolsep{2pt}
  \centering
  \begin{tabular}{lcccccccc}
    \toprule
    \textbf{Dataset} & \textbf{Cora} & \textbf{Cit.} & \textbf{Pub.} & \textbf{CS} & \textbf{Phy.} & \textbf{Pho.} & \textbf{Com.} & \textbf{arX.} \\
    \midrule
    \multicolumn{9}{c}{\textbf{Initial Model Suggestion}} \\
    \midrule
    \textbf{Llama2} & 80.26 & 68.90 & 75.83 & \textbf{89.84} & 90.75 & \textbf{91.35} & 80.27 & 70.95 \\
    \textbf{GPT-4}  & \textbf{80.31} & \textbf{69.20} & \textbf{76.60} & 89.64 & \textbf{92.10} & 91.19 & \textbf{82.20} & \textbf{71.50} \\
    \midrule
    \multicolumn{9}{c}{\textbf{Model Refinement}} \\
    \midrule
    \textbf{Llama2} & 80.96 & 70.38 & 77.40 & 90.19 & 91.47 & 92.09 & 83.69 & 71.63 \\
    \textbf{GPT-4}  & \textbf{81.77} & \textbf{71.00} & \textbf{77.57} & \textbf{90.51} & \textbf{92.61} & \textbf{92.38} & \textbf{84.08} & \textbf{72.02} \\
    \bottomrule
  \end{tabular}
  \vspace{-6px}
\end{table}

\subsection{Case Studies}
\label{ssec:case_study}
\noindent
\textbf{Different LLMs.}
To study the effect of different LLMs as meta-controllers, we use an open-source Llama2-13b to replace the GPT-4 meta-controller.
Table~\ref{tab:llm_comparison_combined} shows the comparative results on the initial model proposal and refinement.
We observed that replacing GPT-4 with Llama2 results in a slight degradation of the performance of recommended models in both settings, and is more consistent in model refinement.
This suggests that GPT-4 is superior at proposing a reasonable
next step from the trajectory and extra knowledge, consistent with the existing study~\citep{yang2023large}.
Notably, because the initial model proposal phrase relies more on the quality of our elicited knowledge, Llama2 can also achieve close performance to GPT-4.


\noindent
\textbf{Lack of Prior Knowledge in LLMs.}
The \textit{inherent} knowledge gap illustrated in Figure~\ref{subfig:design-issue} refers to the lack of prior knowledge in LLMs about the top-performing models of benchmark datasets within the NAS-Bench-Graph~\citep{qin2022bench} model space.
We analyze LLM responses in which the descriptive inputs or graph properties were directly sent to the LLMs for model suggestions.
The results demonstrate that LLMs are only able to suggest commonplace layer connections and operations based on user-provided descriptive inputs or key graph properties.
We have tested all benchmark datasets in NAS-Bench-Graph and found that only two macro lists and three operation lists would be recommended by LLMs regardless of the datasets:
(1) \textbf{Architecture:} [0, 1, 2, 3] and [0, 0, 1, 3]; (2) \textbf{Operations:} ['gcn', 'gat', 'sage', 'skip'], ['gcn', 'gat', 'sage', 'gcn'], and ['gcn', 'gat', 'gin', 'sage'].
After providing tailored knowledge derived from our framework, the LLM-recommended model becomes specialized and shows significant performance improvements.

\noindent
\textbf{Artificial Hallucination in LLMs.}
In Figure~\ref{subfig:semantic-features}, we examine the phenomenon of ``artificial hallucination,'' or the \textit{external} noise issue, in LLMs when comparing the similarities between unseen and benchmark datasets (Section~\ref{sec:methodology_KR}).
Figure~\ref{fig:heatmap} (leftmost) illustrates that the empirically most similar benchmark datasets to PubMed are CS, Physics, and Citeseer, in descending order of similarity. When employing our Graph Understanding method in Section~\ref{ssec:graph_understanding}, which leverages \textit{confident} graph properties, the top three most similar datasets are Citeseer, CS, and Physics (align with the empirical ground truth).
However, the current practice that relies solely on descriptive inputs identifies Cora, Citeseer, and ogbn-arxiv as the top three, which does not align with the empirical ground truth. This discrepancy arises because LLMs overly rely on the shared characteristic of being citation graphs, assuming that citation datasets like PubMed, Cora, Citeseer, and ogbn-arxiv should have similar model design preferences.
This reveals that relying solely on \textit{external} descriptive inputs is insufficient to capture the similarities between datasets, which can overwhelm other pertinent information, leading to inaccurate task association and knowledge retrieval.

\section{Conclusion}
\label{sec:conclusion}
We present DesiGNN, a knowledge-centered framework that demonstrates deep expertise in \textit{proficiently} designing data-aware GNNs and eliminates \textit{external} noise by systematically converting accumulated model design experience into fine-grained meta-knowledge well suited to meta-learning with LLMs.
Leveraging a performance-oriented task similarity grounded in empirically validated graph properties and adaptive knowledge elicitation via LLMs, DesiGNN bridges the gap between literature insights and practical, data-specific model deployment.
DesiGNN integrates graph understanding, effective knowledge retrieval, and strategic model suggestion and refinement under a Bayesian perspective, enabling tailored GNN designs for diverse graphs in a single shot or with a few attempts.
Extensive experiments demonstrate that DesiGNN expedites the design process by delivering top-tier GNN designs within seconds and achieving superior performance at minimal cost.

\begin{acks}
  Lei Chen's work is partially supported by National Key Research and Development Program of China Grant No. 2023YFF0725100, National Science Foundation of China (NSFC) under Grant No. U22B2060, Guangdong-Hong Kong Technology Innovation Joint Funding Scheme Project No. 2024A0505040012, the Hong Kong RGC GRF Project 16213620, RIF Project R6020-19, AOE Project AoE/E-603/18, Theme-based project TRS T41-603/20R, CRF Project C2004-21G, Key Areas Special Project of Guangdong Provincial  Universities 2024ZDZX1006,  Guangdong Province Science and Technology Plan Project 2023A0505030011, Guangzhou municipality big data intelligence key lab, 2023A03J0012, Hong Kong ITC ITF grants MHX/078/21 and PRP/004/22FX, Zhujiang scholar program 2021JC02X170, Microsoft Research Asia Collaborative Research Grant, HKUST-Webank joint research lab, 2023 HKUST Shenzhen-Hong Kong Collaborative Innovation Institute Green Sustainability Special Fund, from Shui On Xintiandi and the InnoSpace GBA, and HKUST(GZ) - CMCC(Guangzhou Branch) Metaverse Joint Innovation Lab under Grant No. P00659. Jiachuan Wang's work is supported in part by JST CREST (JPMJCR22M2). Shimin Di's work is supported by National Science Foundation of China (NSFC) under Grant No. 62506075.
\end{acks}


\bibliographystyle{ACM-Reference-Format}
\balance
\bibliography{refs}

\section*{Ethical Considerations}
\label{sec:societal_impact}
Our work contributes to the growing movement toward high-level automation in artificial intelligence by enabling the proficient, data-aware design of Graph Neural Networks (GNNs) through structured knowledge reuse and language model alignment. As AI systems become more complex and specialized, the ability to automate model design across unseen tasks without human expertise offers transformative potential for a wide range of real-world applications. 
DesiGNN lowers the barrier to applying advanced graph-based machine learning, making it more accessible to scientists, engineers, and practitioners across domains such as biology, social networks, and recommendation systems. By replacing ad hoc manual tuning with systematic and interpretable design strategies, our framework may promote reproducibility, consistency, and scalability in the deployment of machine learning systems.
More broadly, this work aligns with the vision of LLM-powered agents acting as autonomous scientific assistants---able to rapidly translate raw data into effective models---thereby accelerating innovation in scientific discovery, infrastructure optimization, and decision-making. We hope our contribution supports the responsible expansion of AI automation by enabling efficient and expert-level design processes at scale.

However, such high-level automation also warrants caution. While DesiGNN significantly reduces computational cost and expertise requirements, its efficacy may be affected by the quality and representativeness of the knowledge source, as well as the reliability of LLMs' inherent contextual reasoning capabilities.
Additionally, while our method operates at a level removed from direct applications, any downstream deployment of GNNs---e.g., in social network analysis or recommender systems---can have far-reaching societal implications, such as privacy leakage or algorithmic bias, if not handled responsibly.

\end{document}